\documentclass[letterpaper, 10 pt, conference]{ieeeconf}  

\IEEEoverridecommandlockouts                              

\usepackage{graphics} 
\usepackage{epsfig} 
\usepackage{mathptmx} 
\usepackage{times} 
\usepackage{amsmath} 
\usepackage{amssymb}  
\usepackage{mathtools} 
\usepackage{url}
\usepackage{cite}
\usepackage[inkscapelatex=true]{svg}
\usepackage[caption=false,font=footnotesize]{subfig}
\usepackage{fixltx2e}
\usepackage{amsfonts}
\usepackage{booktabs}
\usepackage{siunitx}
\usepackage{comment}
\usepackage{hyperref}

\title{\LARGE \bf

    Diffusion Sequence Models for\\Generative In-Context Meta-Learning of Robot Dynamics
}

\author{Angelo Moroncelli$^{1,2,*}$\thanks{*These authors contributed equally to this work.}, Matteo Rufolo$^{1,2,*}$, Gunes Cagin Aydin$^{3,*}$, Asad Ali Shahid$^{1,3}$, Loris Roveda$^{1,3}$
\thanks{Corresponding author: {\tt loris.roveda@supsi.ch}.}
\thanks{$^{1}$University of Applied Science and Arts of Southern Switzerland, Department of Innovative Technologies, IDSIA-SUPSI, Lugano, Switzerland.}
\thanks{$^{2}$Università della Svizzera Italiana, Faculty of Informatics, Lugano, Switzerland.}
\thanks{$^{3}$Politecnico di Milano, Mechanical Department, Milano, Italy.}}

\begin{document}

\maketitle
\thispagestyle{empty}
\pagestyle{empty}

\begin{abstract}
Accurate modeling of robot dynamics is essential for model-based control, yet remains challenging under distributional shifts and real-time constraints. In this work, we formulate system identification as an in-context meta-learning problem and compare deterministic and generative sequence models for forward dynamics prediction. We take a Transformer-based meta-model, as a strong deterministic baseline, and introduce to this setting two complementary diffusion-based approaches: (i) inpainting diffusion (Diffuser), which learns the joint input–observation distribution, and (ii) conditioned diffusion models (CNN and Transformer), which generate future observations conditioned on control inputs. Through large-scale randomized simulations, we analyze performance across in-distribution and out-of-distribution regimes, as well as computational trade-offs relevant for control. We show that diffusion models significantly improve robustness under distribution shift, with inpainting diffusion achieving the best performance in our experiments. Finally, we demonstrate that warm-started sampling enables diffusion models to operate within real-time constraints, making them viable for control applications. These results highlight generative meta-models as a promising direction for robust system identification in robotics.
\end{abstract}

\section{Introduction}
\label{introduction}


Accurate modeling of system dynamics lies at the core of robot control~\cite{paramidrobots1985}, underpinning applications including model predictive control (MPC)~\cite{5153127} and model-based reinforcement learning~\cite{ramesh2023physicsinformedmodelbasedreinforcementlearning}. However, accurate and reliable modeling of real-world robotic systems remains inherently challenging, as classical physics-based approaches often struggle to fully capture the complexity of real-world dynamics. Data-driven approaches offer an appealing alternative by directly learning robot behavior from observations~\cite{chi2024diffusionpolicyvisuomotorpolicy,Moroncelli2024TheDO}. In particular, black-box models approximate system dynamics as a function of input-output trajectories without requiring explicit parameterization. Despite their flexibility, such methods often suffer from poor generalization, high data requirements, and limited robustness under distributional shifts~\cite{Ai2025LearningDynamicsReview}.

Within this landscape, learning-based approaches to robot control can be broadly categorized into three paradigms: (i) policy learning methods, which directly map observations to inputs~\cite{chi2024diffusionpolicyvisuomotorpolicy}; (ii) world models, which learn latent representations optimized for planning and control~\cite{hansen2024tdmpc2}; and (iii) explicit dynamics models, which predict future system observations~\cite{giacomuzzos2023blackbox,bazzi2024robomorphincontextmetalearningrobot}. Among these, explicit dynamics models offer a natural interface with classical control techniques, but their effectiveness critically depends on accurate system identification, which remains challenging in practice.

In this work, we approach the dynamics modeling problem through the lens of meta-learning. We do this by adopting a \textbf{black-box meta-modeling} framework for dynamics, casting system identification as an in-context learning problem. This paradigm was initially proposed by learning a meta-model that represents a class of dynamical systems harnessing the power of Transformers~\cite{forgione2023modelsclassmodelsincontext,vaswani2023attentionneed} and subsequently extended in future works~\cite{piga2024syntheticdatagenerationidentification,rufolo2025distributionally}. More recently, it has been successfully scaled to high-dimensional robotic manipulation tasks \cite{bazzi2024robomorphincontextmetalearningrobot,elseiagy2026data}. The core premise relies on the in-context learning capabilities of modern neural architectures. Rather than optimizing a separate neural network for every distinct system, the meta-model learns the governing rules of entire classes of dynamical systems from contextual input-output trajectories. This framework provides a powerful, data-driven mechanism for generalization across similar systems by effectively ``learning to learn'' \cite{vanschoren2018meta}. Transformer-based meta-models such as RoboMorph~\cite{bazzi2024robomorphincontextmetalearningrobot}, provide a strong deterministic baseline for explicit robot dynamics modeling via in-context learning. However, these models inherently lack the capacity for rigorous uncertainty quantification and suffer from severe performance degradation when exposed to distributional shifts.

\begin{figure*}[t]
\centering
    \includegraphics[width=\linewidth]{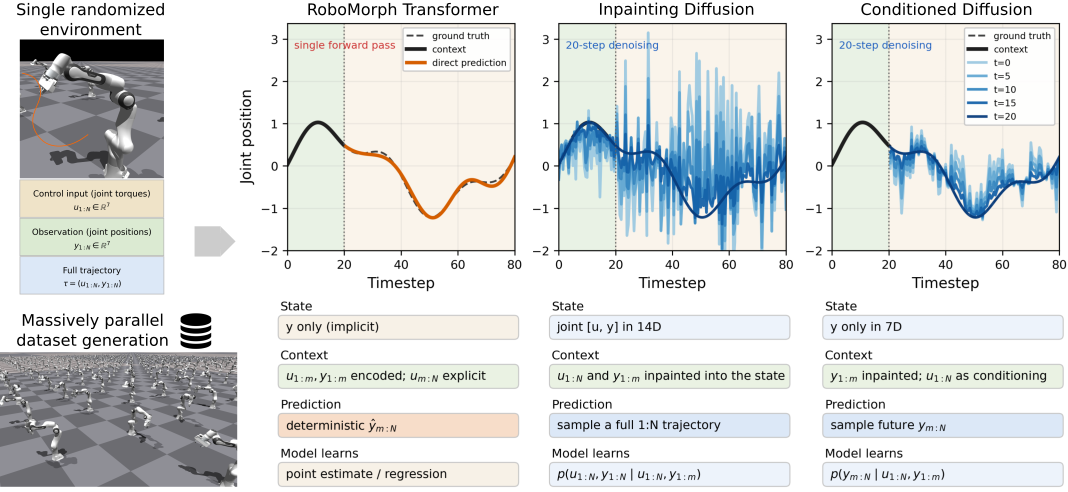}
    \caption{\textbf{Comparison of sequence models.} Diffusion models (\textit{Inpainting} and \textit{Conditioned}) are contrasted with a deterministic Transformer (\textit{RoboMorph}). Inpainting Diffusion learns the full trajectory distribution over $y_{1:N}$ from context (the green area). Conditioned Diffusion models system dynamics by generating $y_{m:N}$ conditioned on $u_{1:N}$, producing future trajectories under input conditioning with smoother denoising updates. Diffusion models require multiple iterative steps (e.g., 20 in this example) to progressively denoise an initial random distribution, refining predictions over time. In contrast, the deterministic Transformer predicts $y_{m:N}$ directly from $u_{m:N}$ and context in a single forward pass.
}
\label{fig:architectures}
\end{figure*}

While recent advances in diffusion models have achieved impressive results in policy generation~\cite{chi2024diffusionpolicyvisuomotorpolicy} and trajectory planning~\cite{janner2022planningdiffusionflexiblebehavior}, and Denoising Diffusion Probabilistic Models (DDPMs)~\cite{ho2020denoisingdiffusionprobabilisticmodels} have recently emerged as stable generative frameworks capable of modeling multi-modal distributions~\cite{wang2023incontextlearningunlockeddiffusion}; these approaches typically bypass explicit modeling of system dynamics. Despite their success in policy learning, their application to explicit dynamics estimation remains largely unexplored, particularly in meta-learning settings~\cite{bazzi2024robomorphincontextmetalearningrobot,meng2024metapnn}. Consequently, a gap remains between advances in generative modeling and the requirements of system identification for control~\cite{Ai2025LearningDynamicsReview}.

In this work, as shown in Fig.~\ref{fig:architectures}, we propose two diffusion-based formulations for system identification: \textbf{inpainting diffusion} (\textit{Diffuser})~\cite{janner2022planningdiffusionflexiblebehavior}, which models the full input–observation trajectory, and \textbf{conditioned diffusion}~\cite{chi2024diffusionpolicyvisuomotorpolicy}, which predicts future observations conditioned on control inputs using CNN (\textit{CDCNN}) or Transformer (\textit{CDT}) backbones. By systematically evaluating these novel probabilistic frameworks against established deterministic baselines, this paper seeks to answer the following critical research questions:
\begin{itemize}
    \item Can generative models improve robustness and generalization in system identification?
    \item How do different diffusion formulations compare to deterministic architectures in modeling complex robot dynamics?
    \item What are the trade-offs between prediction accuracy and computational cost in control-oriented settings?
\end{itemize}

To address these questions, we develop a unified experimental framework for meta-learned robot dynamics across diverse randomized systems and excitation signals. We introduce diffusion-based generative models for in-context dynamics learning, explicitly modeling forward dynamics, unlike diffusion policy methods. Compared to deterministic meta-models, our approach captures the trajectory distribution, improving robustness under distributional shift while remaining compatible with real-time control via warm-starting. Our main contributions are:
\begin{itemize}
    \item We extend in-context meta-learning for robot dynamics to a large-scale randomized simulation setting.
    \item We introduce diffusion-based generative models (inpainting and conditioned) for dynamics meta-modeling and compare them to a deterministic Transformer baseline (RoboMorph) under distribution shift.
    \item We show that warm-started conditioned diffusion enables real-time, control-compatible inference while maintaining strong robustness.
\end{itemize}

The remainder of this paper is organized as follows. Section~\ref{problem_description} introduces the problem formulation and proposed models, Section~\ref{benchmarks} presents the experimental evaluation, and Section~\ref{conclusion} concludes the work. Models checkpoints and datasets are publicly available.

\section{Problem Description}
\label{problem_description}

In this section, we formalize our meta-learning framework and motivate our architectural choices, domain randomization strategy, and training procedures.

Classical modeling of robot dynamics relies on deriving a faithful mathematical representation of a physical plant. Formally, let $\mathcal{S}$ denote a specific physical \textit{system} (e.g., a robotic manipulator with exact, fixed physical parameters) drawn from a broader \textit{system class} $\mathcal{C}$ of similar systems, which represents the family of all such systems under varying physical conditions (e.g. different physical parameters). 
When exact prior knowledge about the physical parameterization of $\mathcal{S}$ is unavailable, system identification relies on black-box modeling. This approach is agnostic to the underlying physical equations, instead approximating the true system dynamics via a parameterized function approximator $g(x, \theta)$. To optimize $\theta$, the model is trained on a trajectory dataset $\mathcal{D} = (u_{1:N}, y_{1:N})$, which comprises a finite sequence of control inputs and corresponding system observations generated by exciting the specific physical system $\mathcal{S}$. Depending on the required expressiveness, this model can range from classical linear projections over a set of basis functions to highly non-linear, high-dimensional neural network architectures\cite{forgione2021dynonet}.

\subsection{Model-Free Black-Box Meta-Models}
\label{methodology:models}

While traditional black-box models are trained to identify a single, isolated dynamical system, recent advancements have expanded this paradigm to model entire classes of systems \cite{forgione2023modelsclassmodelsincontext}. This is achieved by framing system identification as an \textit{meta-learning} problem. In this framework, a neural meta-model $\mathcal{M}_\phi$ is trained directly over a broad trajectory distribution $p(\mathcal{D})$, which jointly encapsulates the variations in underlying physical systems and the corresponding control excitations.

For each sampled trajectory ${D} \sim p(\mathcal{D})$, we partition the data into a context window of length $m$, and a prediction horizon from $m$ to $N$. The context, denoted as $D_{ctx} = (u_{1:m}, y_{1:m})$, provides the necessary historical information to implicitly identify the specific system dynamics. The meta-model is then tasked with predicting the future system response $y_{m:N}$, given on both the context and the future control inputs $u_{m:N}$:
\begin{equation}
    \hat{y}_{m:N} = \mathcal{M}_\phi\left(u_{m:N}, {D}_{ctx}\right).
\end{equation}
The optimal model parameters $\phi^*$ are obtained by minimizing the expected prediction loss (e.g., Mean Squared Error (MSE)) across the entire trajectory space:
\begin{equation}
\label{eq:mse_Loss}
    \phi^* = \arg\min_\phi \mathbb{E}_{{D} \sim p(\mathcal{D})} \left[ \left\| y_{m:N} - \hat{y}_{m:N} \right\|_2^2 \right].
\end{equation}
Transformers, inherently designed for sequence-to-sequence mapping and in-context conditioning \cite{52580}, serve as a natural architectural baseline for this meta-modeling task. 

\subsection{Deterministic vs. Generative Inference}
\label{methodology:model-inference}

Standard neural architectures trained via the deterministic objective above, regress a single point estimate.

To account for complex model and data-borne uncertainties, the meta-model must be formulated probabilistically to explicitly learn the conditional distribution of the system trajectories. This is achieved by transitioning from deterministic point estimation to a probabilistic meta-model $p_\phi$, which maximizes the expected log-likelihood over the task distribution:
\begin{equation}
    \phi^* = \arg\max_\phi \mathbb{E}_{\mathcal{D} \sim p(\mathcal{D})} \left[ \log p_\phi \left( y_{m:N} \mid u_{m:N}, D_{ctx} \right) \right].
\end{equation}
While this formulation provides a principled measure of uncertainty, standard implementations typically restrict the predictive estimation to uni-modal probability distributions. For a comprehensive derivation of this probabilistic framework within a system identification context, we refer the reader to previous work~\cite{rufolo2025enhanced}.

While standard architectures can be extended into this generative framework (e.g., via Variational Autoencoders), doing so typically requires explicitly defining complex, rigid priors over a reduced latent space, which can overly restrict expressiveness when modeling high-dimensional robot dynamics. To robustly parameterize $p_\phi$ without restrictive latent assumptions, DDPM\cite{ho2020denoisingdiffusionprobabilisticmodels} have proven highly effective. A DDPM defines a forward Markov chain that gradually corrupts the true future trajectory, denoted as $y^0 = y_{m:N}$, with Gaussian noise over $T$ steps, alongside a parameterized reverse process that learns to iteratively denoise it. Let $y^t$ denote the noisy trajectory at diffusion step $t \in \{1, \dots, T\}$. The forward noise-addition process is governed by a fixed variance schedule, yielding a sequence of increasingly noisy observations.

To reverse this process, the generative meta-model $\mathcal{M}_\phi$ is trained to predict the exact injected noise $\epsilon_\phi$, conditioned on the current noisy observation $y^t$, the future inputs $u_{m:N}$, and the past context $D_{ctx} = (u_{1:m}, y_{1:m})$. During inference, starting from pure Gaussian noise $y^T \sim \mathcal{N}(\mathbf{0}, \mathbf{I})$, the future trajectory is iteratively reconstructed by sampling from the learned posterior $p_\phi^*$. At each step $t$, the predicted noise $\epsilon_\phi$ is subtracted from $y^t$ (scaled by the predefined diffusion scheduling parameters) alongside a stochastic Gaussian injection, progressively resolving the true trajectory $y^0$.

To optimize $\mathcal{M}_\phi$, we minimize the discrepancy between the true injected Gaussian noise $\epsilon \sim \mathcal{N}(\mathbf{0}, \mathbf{I})$ and the predicted noise $\epsilon_\phi$ using the standard simplified variational bound \cite{ho2020denoisingdiffusionprobabilisticmodels}. Unlike deterministic baselines (e.g., RoboMorph), which optimize a direct trajectory error $\mathcal{L}_{\text{det}} = \|y_{m:N} - \hat{y}_{m:N}\|_2^2$, our diffusion models are trained using a weighted MSE loss applied at randomly sampled timesteps $t$:
\begin{equation}
\label{eq:diffusion_Loss}
    \mathcal{L}_{\text{diff}} = \mathbb{E}_{t, y^0, \epsilon} \left[ \left\| W \odot \left(\epsilon - \mathcal{M}_\phi(y^t, t, u_{m:N}, D_{ctx}) \right) \right\|_2^2 \right].
\end{equation}
Here, $W$ represents a constant weight mask applied over the joint input-observation space. Because future control inputs $u_{m:N}$ are perfectly known deterministically, the mask is set to $W = [w_u, w_y] = [1, 3]$ for the Diffuser architecture. This selectively penalizes observation reconstruction errors, actively forcing the network to prioritize learning the unknown system dynamics rather than reconstructing the given control sequences. For all other diffusion models, a standard unweighted mask $W = [1, 1]$ is utilized.

Fundamentally, unconditional diffusion models act as pure generative priors; they blindly sample from the learned data distribution without regard for specific environmental constraints or target outcomes. To effectively steer the generative process toward a desirable, dynamically valid trajectory, the sampling must be explicitly guided via goal-conditioned loss functions, architectural conditioning, or inpainting. The specifics of these structural formulations are detailed in the subsequent section.

\subsection{Neural Architectures}
\label{methodology:neural-architectures}

We consider sequence models along two dimensions: (i) \emph{deterministic vs. generative inference}, and (ii) \emph{inpainting vs. conditional trajectory modeling}, as illustrated in Fig.~\ref{fig:architectures}. This framing enables a unified analysis of expressiveness, robustness, and computational efficiency in meta-learned dynamics.

We adopt standard architectures in robotics, namely Transformers~\cite{vaswani2023attentionneed} and CNNs~\cite{ronneberger2015unetconvolutionalnetworksbiomedical}, instantiated within the meta-learning framework described above. 

\paragraph{\textbf{Transformer (RoboMorph)}}
RoboMorph~\cite{bazzi2024robomorphincontextmetalearningrobot} serves as a deterministic baseline based on an encoder--decoder Transformer~\cite{vaswani2023attentionneed}. The context $(u_{1:m}, y_{1:m})$ is encoded and cross-attended with future inputs $u_{m:N}$ to predict $\hat{y}_{m:N}$. While effective in simple in-distribution settings, it performs deterministic regression, approximating $\mathbb{E}[y_{m:N} \mid u_{m:N}, D_{ctx}]$. Consequently, it yields a single point estimate that fails to quantify epistemic or aleatoric uncertainty, fundamentally limiting its applicability in safety-critical robotic tasks where uncertainty is required.

\paragraph{\textbf{Inpainting Diffusion (Diffuser)}}
To address the above limitation, we introduce a generative approach based on diffusion models. The Diffuser~\cite{janner2022planningdiffusionflexiblebehavior} models the \emph{joint distribution} over input--observation trajectories using a DDPM~\cite{ho2020denoisingdiffusionprobabilisticmodels} with a U-Net backbone~\cite{ronneberger2015unetconvolutionalnetworksbiomedical}. Known values $(u_{1:N}, y_{1:m})$ are enforced via inpainting at each denoising step. By modeling $p(u,y \mid u_{1:N}, y_{1:m})$, Diffuser captures rich input--observation correlations, yielding expressive and multi-modal predictions with strong robustness, especially out-of-distribution. This increased expressiveness, however, comes with higher computational cost and greater sensitivity to truncated (warm-started) inference.

\paragraph{\textbf{Conditioned Diffusion (CDCNN and CDT)}}
We also propose conditioned diffusion, which models $p(y_{m:N} \mid u_{1:N}, y_{1:m})$, reducing the complexity of the generative task. Control inputs are injected via FiLM conditioning~\cite{perez2017filmvisualreasoninggeneral}, following recent approaches for generative policies~\cite{chi2024diffusionpolicyvisuomotorpolicy}. 

We instantiate this formulation with both CNN (CDCNN) and Transformer (CDT) backbones. CNN-based models enforce local temporal smoothness through convolutional filtering, producing physically coherent trajectories, while Transformer-based models capture long-range dependencies but may introduce higher-frequency oscillations due to the lack of local inductive bias. Despite these differences, both variants retain multi-modal expressiveness and benefit from more efficient and stable inference compared to Diffuser.

Overall, these architectures define a clear trade-off. Deterministic Transformers are fast but brittle under distribution shifts. Inpainting diffusion maximizes expressiveness and robustness by modeling full state distributions, but is computationally heavier. Conditioned diffusion provides an effective middle ground, achieving strong robustness in trajectory prediction of $y_{m:N}$ with significantly improved efficiency.

\subsection{Datasets}
We train our black-box meta-models over a wide range of geometric configurations and dynamical parameters of the Franka Emika Panda, using nominal values from~\cite{8772145}. System parameters are randomized and $3\times10^5$--$10^6$ robots are simulated in parallel using IsaacGym~\cite{makoviychuk2021isaac}.


For feedforward excitation, joint torques are generated using chirp (CH) and multi-sinusoidal (MS) signals. The chirp excitation is defined as $u_{CH}(t)=A\cos\!\big(\omega_1(1+\tfrac{1}{4}\cos(\omega_2 t))t+\phi\big)$, where $A$ and $\phi$ are randomized amplitude and phase, and $\omega_1=2\pi f_1$, $\omega_2=2\pi f_2$ control the time-varying frequency. The multi-sinusoidal input is defined as $u_{MS}(t)=\sum_{k=0}^{3} A_k\,\psi_k(\omega_k t)$, where $\psi_k\in\{\sin,\cos\}$, $\omega_k\in\{\omega_0,1.5\omega_0,2\omega_0,3\omega_0\}$, $\omega_0=2\pi f_0$, and $A_k$ are randomized amplitudes.

Each dataset consists of 7-dimensional input sequences $u_{1:N}$ (joint torques) and 7-dimensional observations $y_{1:N}$ (joint observations), with straightforward extensions to higher-dimensional representations including Cartesian and end-effector dynamics.

\label{methodology:datasets}
\begin{table}[t!] 
\caption{Simulation Parameters Randomization}
\centering
\begin{tabular}{|c|c|c|} \toprule
     Dataset and Signal& $A [Nm]$& $f [Hz]$\\ \midrule\midrule
     $D_{1}$ -- $CH$&  $[-4,4]$& $0.3$\\\midrule      
     $D_{2}$ -- $CH$&  $[-4,4]$& $[0.2,0.4]$\\ 
     $D_{3}$ -- $CH$&  $[-4,4]$& $[0.2,0.6]$\\ 
     $D_{4}$ -- $CH$&  $[-4,4]$& $[0.1,0.7]$\\ \midrule
     $D_{1}$ -- $MS$&  $[-30f,30f]$& $0.15$\\\midrule    
     $D_{2}$ -- $MS$&  $[-30f,30f]$& $[0.05,0.15]$\\ 
     $D_{3}$ -- $MS$&  $[-30f,30f]$& $[0.05,0.25]$\\ 
     $D_{4}$ -- $MS$&  $[-30f,30f]$& $[0.01,0.30]$\\ \midrule
\end{tabular}
\label{tab:randomization}
\end{table}
The inputs, initial conditions, and dynamical parameters are randomized; the corresponding amplitude and frequency ranges for each training dataset are summarized in Table~\ref{tab:randomization}, from which all signal parameters (i.e., $A$, $f_1$, $f_2$ for CH, and $A_k$, $f_0$ for MS) are sampled.

\subsection{Training Procedures}
\label{methodology:training}
To optimize the meta-models, we minimize the empirical formulations of the expected losses defined in Equations~\ref{eq:mse_Loss} and~\ref{eq:diffusion_Loss}. Because the true mathematical expectations over the system distribution are analytically intractable, they are approximated via Monte Carlo sampling across discrete mini-batches of size $64$. Specifically, all architectures are trained over $3.5 \times 10^6$ fully randomized robotic trajectories, spanning $10$ full training epochs each (as illustrated in Fig.~\ref{fig:architectures}). Each trajectory consists of $N = 400$ time steps, which are partitioned into a $320$-step context window and an $80$-step prediction horizon. Optimization is performed using the Adam optimizer coupled with a cosine annealing learning rate scheduler, which gracefully decays the learning rate to one-tenth of its initial value by the end of training. The initial learning rate is set to $6 \times 10^{-4}$ for RoboMorph and Diffuser, and $1 \times 10^{-4}$ for the conditioned architectures (CDCNN and CDT). All training procedures were executed on a single NVIDIA A100 GPU.

Regarding architectural configurations, both RoboMorph and CDT employ $12$ MLP layers, $8$ attention heads, and an embedding dimension of $384$, directly adopting the optimized hyperparameters established in~\cite{bazzi2024robomorphincontextmetalearningrobot}. Conversely, the fully convolutional architectures (Diffuser and CDCNN) are configured with $128$ initial convolution layers and $3$ downsampling/upsampling steps. For generative inference, all three diffusion models (Diffuser, CDCNN, and CDT) utilize a $100$-step denoising schedule and are explicitly optimized to predict the injected Gaussian noise. The hyperparameters governing the diffusion processes were selected following a systematic ablation study to ensure optimal predictive performance. Under these configurations, the total offline training times were approximately $4.5$ hours for RoboMorph, $3.7$ hours for Diffuser, $2.0$ hours for CDCNN, and $6.7$ hours for CDT.

\begin{figure*}[t!]
\centering
\includegraphics[width=0.6\linewidth]{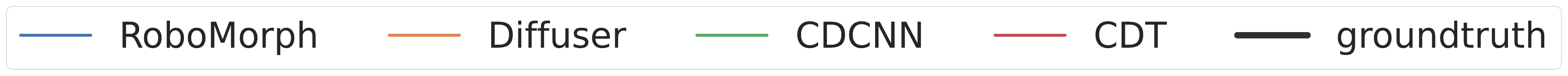}
\centering
\includegraphics[height=2.0in]{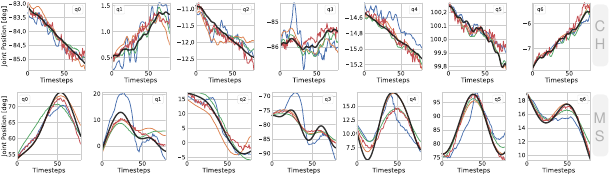}
\caption{Meta-modeling of joint dynamics for chirp and multi-sinusoidal joint torques with $f_{CH}=1.0Hz$, $f_{MS}=0.45Hz$ master frequencies (OOD), trained on $D_1$. The results indicate one prediction per signal among $100$ randomized scenarios. Diffusion-based architectures seem to better capture global complexity, particularly in OOD signals, such as in this example.}
\label{fig:ff_simulation}
\end{figure*}

\section{Simulation Results and Analysis}
\label{benchmarks}

In this section, we evaluate the performance of the proposed framework across diverse simulation scenarios. All experiments are evaluated in both in-distribution (ID) and out-of-distribution (OOD) settings. We first assess performance by evaluating the accuracy and adaptability of different architectures across datasets. We then analyze inference time and compare the models from a closed-loop control perspective.

\subsection{Forward Dynamics Meta-Model Performance}

Here, we consider a wide range of CH and MS signals. CH signals are generally easier to model, as at both low and high frequencies they tend to converge to stationary or monotonically increasing trajectories, which are relatively simple for neural networks to learn, as shown in Fig.~\ref{fig:ff_simulation}. In contrast, MS signals do not reach a steady-state plateau. This persistent transient excitation becomes more pronounced at higher frequencies (around $0.3$--$0.5$~Hz), resulting in jagged and rapidly varying dynamics that pose a greater predictive challenge. Fig.~\ref{fig:ff_simulation} illustrates this behavior, highlighting how the continuous superposition of sinusoidal components complicates prediction across the entire trajectory.

By meta-modeling the forward dynamics, we are able to accurately predict most of the challenging signals in our dataset with errors bounded by $3.5$ degrees in joint space as Fig.~\ref{fig:ff_simulation} shows. Nevertheless, not all architectures behave the same.  Trajectories predicted by CNN-based diffusion models are inherently smooth, whereas RoboMorph and CDT, are highly prone to high-frequency oscillatory estimations. This behavior aligns with the fundamental architectural differences between the models. While the Transformer decoder employs causal attention to enforce strict temporal directionality, it still relies on a global self-attention mechanism over the past sequence. Because it lacks a strict inductive bias for local temporal continuity, adjacent time steps can fluctuate independently. Conversely, CNNs possess a strong local inductive bias; their convolutional kernels act as local filters over sliding temporal windows. This explicitly ties adjacent observations together, enforcing temporal coherence and yielding naturally smooth trajectories. In applications where Transformer-induced high-frequency jitter is problematic, applying a standard low-pass denoising filter as a post-processing step could effectively recovers signal continuity.


\begin{figure}[t!]
\centering
\includegraphics[width=0.8\linewidth]{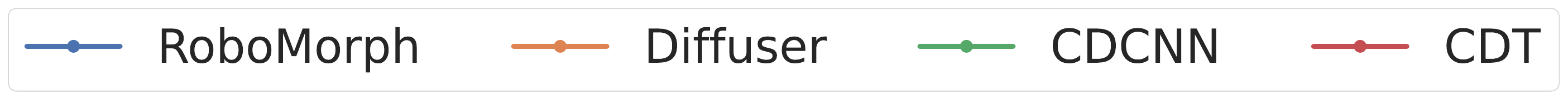}
\centering
\includegraphics[height=1.65in]{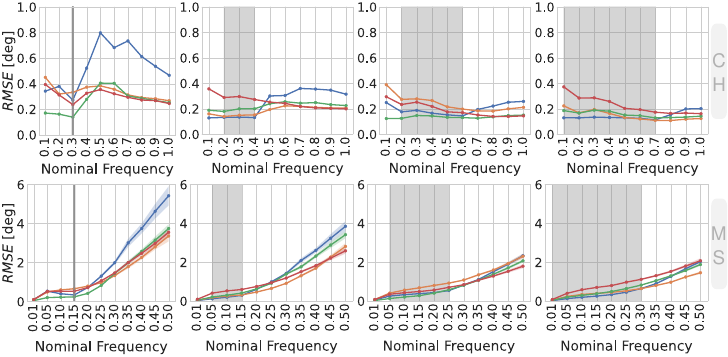}
\caption{By selectively covering parts of the domain, it is possible to meta-learn the class of frequency response. For chirp signals of different randomization: $f_{D_1}=0.30$, $f_{D_2}=[0.02,0.4]$, $f_{D_3}=[0.2,0.6]$, $f_{D_4}=[0.1,0.7]$ and sinusoidal signals of different randomization bounds $f_{D_1}=0.15$, $f_{D_2}=[0.05,0.15]$, $f_{D_3}=[0.05,0.25]$, $f_{D_4}=[0.01,0.30]$ the ID and OOD regions shift which effectively enlarges the accuracy of the predictive domain. This improvement on the edges of the domain minimally affects the accuracy observed for the centering points.}
\label{fig:RMSE_performances}
\end{figure}

Fig.~\ref{fig:RMSE_performances} illustrates the predictive performance of the evaluated architectures across varying nominal frequencies, corresponding to the parameter ranges detailed in Table~\ref{tab:randomization}. The shaded gray regions denote the ID training domains, while the white regions correspond to OOD scenarios. A primary observation is that the RoboMorph experiences severe performance degradation in OOD regimes for both chirp and sinusoidal signals. In contrast, the diffusion-based models exhibit significantly greater robustness, maintaining stable accuracy even well outside the training distribution. At lower frequencies in the MS tasks, RoboMorph performs on par with the other models, with no statistically significant gap in accuracy. Because all evaluated architectures possess sufficient capacity to model slow, quasi-stationary dynamics, no specific architectural advantage is evident in these low-frequency regimes. However, at higher frequencies, most notably in the MS tasks, RoboMorph's performance deteriorates sharply compared to the diffusion-based models. This disparity highlights the advantage of the diffusion formulations, whose generative modeling capabilities provide superior generalization and robustness in challenging OOD scenarios.

Furthermore, expanding the training dataset from a narrow to a broader frequency domain substantially enhances the predictive accuracy of the deterministic RoboMorph baseline. This behavior highlights a core limitation of standard deterministic meta-learning: robust OOD generalization requires exposing the model to exhaustive dynamical variations, otherwise the framework collapses into narrow, task-specific memorization~\cite{kirschgeneral}. Conversely, this degradation is far less pronounced in the diffusion-based architectures. By explicitly modeling the generative probability distribution rather than regressing a single point estimate, diffusion models inherently capture broader dynamical representations. Consequently, they maintain robust predictive performance even when subjected to limited training diversity. Overall, the diffusion-based architectures consistently outperform the classic RoboMorph across varied scenarios, including several ID cases.

Beyond predictive accuracy, it is necessary to consider the computational trade-offs. In our experiments, RoboMorph demanded significantly longer offline training times, to achieve convergence, compared to the fully convolutional architectures. However, this front-loaded computational cost is ultimately offset during deployment, as the deterministic baseline operates at substantially faster online inference than the iterative denoising processes required by the generative models. As shown in Fig.~\ref{fig:warmstarttimes}, faster inference allows Transformer-based models to be readily deployed in real-time control scenarios~\cite{5153127}, and sampling techniques can be further employed to minimize the inference latency gap.

\subsection{Inference Comparison in Control Perspective}

Inference latency in diffusion models is inherently high, as they require a full forward pass at every denoising timestep. Naively reducing the number of timesteps during training degrades prediction accuracy. Instead, we adopt a more flexible strategy: we train on a dense diffusion schedule and accelerate inference via \emph{warm-starting}~\cite{janner2022planningdiffusionflexiblebehavior}. The reverse process is initialized from a prior trajectory estimate (e.g., the solution from the previous control step) rather than from pure noise, so that only the final fraction of denoising steps must be executed. This substantially reduces latency while preserving most of the predictive performance, making diffusion architectures compatible with real‑time receding‑horizon control.

\begin{figure}[t!]
\centering
\includegraphics[width=0.6\linewidth]{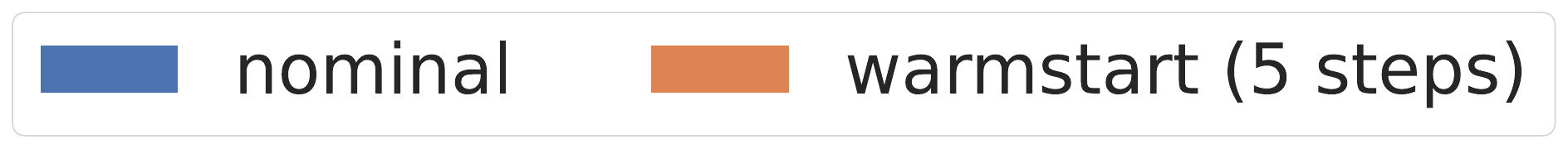}
\includegraphics[height=1.65in]{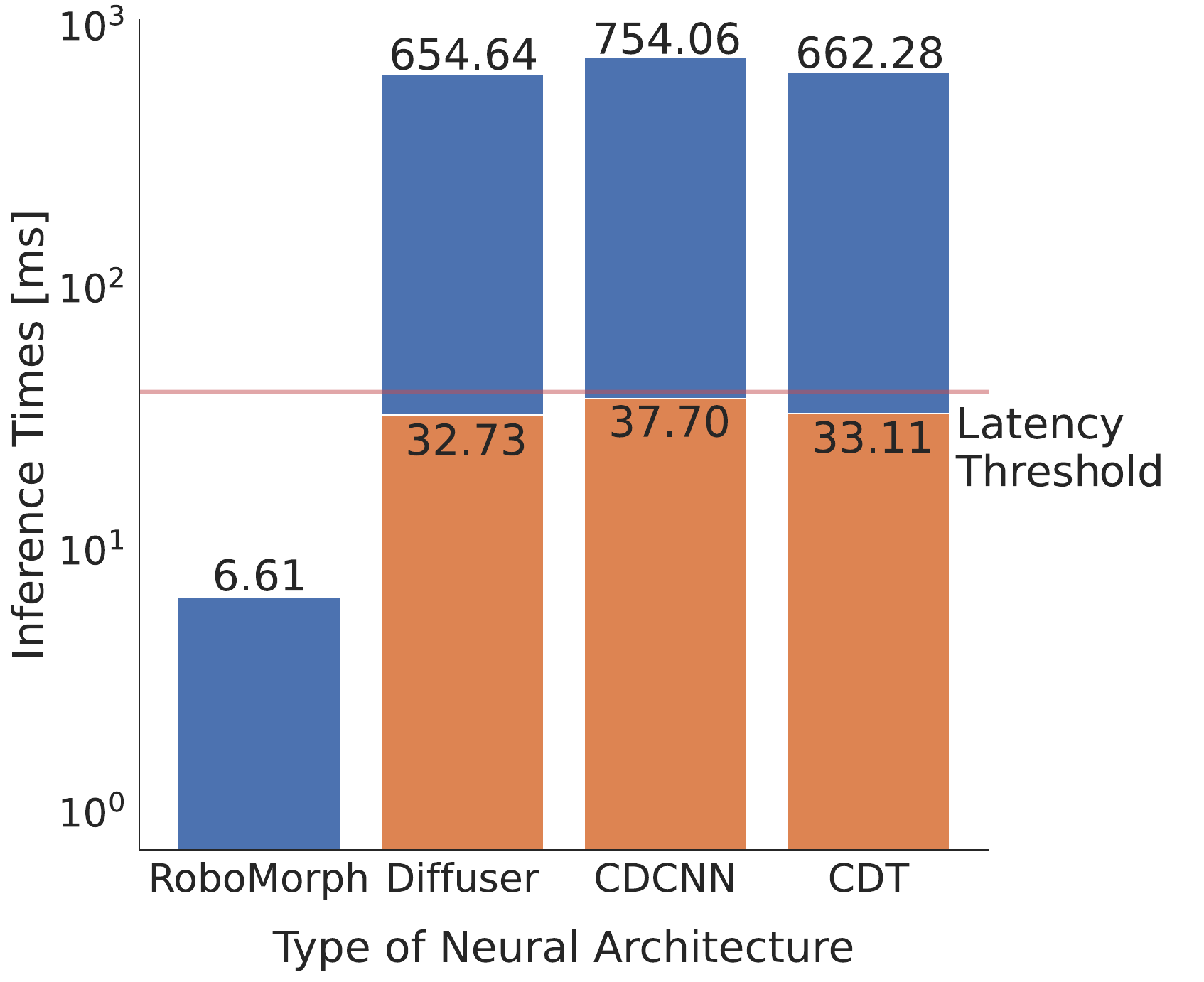}
\caption{Diffusion-based inference is about $2$ order of magnitude larger than non-diffusion-based inference and is a bottleneck of predictive latency. This can be alleviated by warm-starting diffusion models from past trajectories, bringing the inference time to about $40ms$ with $5$ denoising steps.}
\label{fig:warmstarttimes}
\end{figure}

\begin{figure}[t!]
\centering
\includegraphics[width=0.8\linewidth]{images/legend3.pdf}
\includegraphics[height=1.7in]{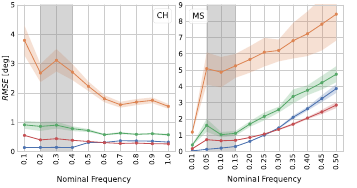}
\caption{As an example, we focus on architectures trained on $D_2$ for chirp and sinusoidal signals with $5$ diffusion steps corresponding to about $40ms$ inference time: Transformer-based models sustain warm-started trajectories relatively well whereas convolution-based models suffer drastically. Among Transformer-based models, diffusion processes provide more versatile representations in modeling high frequency responses whereas for low frequency responses this versatility is practically insignificant.}
\label{fig:warmstartrmse}
\end{figure}

Fig.~\ref{fig:warmstarttimes} reports the resulting inference times. We impose a conservative inference latency threshold of approximately $40$ ms, which corresponds to about $5\%$ of the original diffusion steps (5 warm-started iterations in our implementation). The RMSE degradation induced by this truncation is shown in Fig.~\ref{fig:warmstartrmse}. For this analysis, we focus exclusively on models trained on dataset $D_2$. This choice is empirically justified by the results in Fig.~\ref{fig:RMSE_performances}, which demonstrate that expanding the training distribution to $D_3$ yields only marginal accuracy improvements, indicating that the generalization performance has largely plateaued. The CNN-based diffusion architectures are the most affected, particularly the inpainting variant, suggesting a stronger dependence on the full denoising chain. In contrast, the CDT remains largely insensitive to warm-starting, retaining superior OOD performance while only underperforming the RoboMorph in the ID region.

From a control perspective, Transformer-based models are naturally well suited to high-frequency operation once their inference pipeline is optimized, and we regard low-level control implementations with strict latency constraints as a straightforward extension. In such settings, diffusion models should be systematically warm-started to comply with tight real-time budgets. On the other hand, diffusion models, being inherently multi-modal over the trajectory space, are particularly well suited to robotics settings characterized by highly complex and diverse trajectories, where accurate, long-horizon, and explicit dynamic modeling is desirable; they are especially attractive when OOD generalization is critical.

\section{Conclusion}
\label{conclusion}

In this work, we studied black-box meta-modeling for robotic system identification through a systematic comparison of deterministic and generative sequence models. By casting dynamics learning as an in-context meta-learning problem, we evaluated how implementation choices impact accuracy, robustness, and control-oriented deployment.

Our results highlight three main findings. First, deterministic Transformer-based models such as RoboMorph perform well in-distribution settings but degrade under distributional shifts, especially for complex multi-frequency dynamics. Second, diffusion-based models significantly improve robustness by modeling trajectory distributions; among them, the inpainting joint formulation (Diffuser) achieves the best performance in our experiments due to its richer input--observation representation. Third, conditioned diffusion provides the best trade-off between performance and efficiency, retaining strong robustness while enabling warm-started inference compatible with real-time control.

Overall, the choice between inpainting and conditioned diffusion governs the balance between expressiveness and deployability. While inpainting diffusion is the most expressive, conditioned diffusion emerges as the most practical solution for control-oriented applications.

Future work will focus on real-world validation and integration within MPC pipelines, enabling data-driven receding-horizon control on physical systems. Additionally, exploring mechanistic interpretability to extract physically meaningful parameters from learned models offers a promising direction to bridge deep meta-learning with classical system identification.





\section*{Acknowledgment}
This research was funded by the Swiss National Science Foundation (SNSF), through the project "VR-HRC: Virtual Reality-based Multi-Human-Multi-Robot Collaboration in Industrial Environments", grant number 235540.


\bibliographystyle{IEEEtran}
\bibliography{references}

@inproceedings{janner2022planningdiffusionflexiblebehavior,
  title = {Planning with Diffusion for Flexible Behavior Synthesis},
  author = {Michael Janner and Yilun Du and Joshua Tenenbaum and Sergey Levine},
  booktitle = {International Conference on Machine Learning},
  year = {2022},
}

@article{chi2024diffusionpolicyvisuomotorpolicy,
  title={Diffusion policy: Visuomotor policy learning via action diffusion},
  author={Chi, Cheng and Xu, Zhenjia and Feng, Siyuan and Cousineau, Eric and Du, Yilun and Burchfiel, Benjamin and Tedrake, Russ and Song, Shuran},
  journal={The International Journal of Robotics Research},
  volume={44},
  number={10-11},
  pages={1684--1704},
  year={2025},
  publisher={Sage Publications Sage UK: London, England}
}

@article{forgione2023modelsclassmodelsincontext,
  title={From system models to class models: An in-context learning paradigm},
  author={Forgione, Marco and Pura, Filippo and Piga, Dario},
  journal={IEEE Control Systems Letters},
  volume={7},
  pages={3513--3518},
  year={2023},
  publisher={IEEE}
}

@inproceedings{rufolo2025enhanced,
  title={Enhanced Transformer architecture for in-context learning of dynamical systems},
  author={Rufolo, Matteo and Piga, Dario and Maroni, Gabriele and Forgione, Marco},
  booktitle={2025 European Control Conference (ECC)},
  pages={819--824},
  year={2025},
  organization={IEEE}
}

@article{rufolo2025distributionally,
  title={Distributionally robust minimization in meta-learning for system identification},
  author={Rufolo, Matteo and Piga, Dario and Forgione, Marco},
  journal={IEEE Control Systems Letters},
  year={2025},
  publisher={IEEE}
}

@article{vaswani2023attentionneed,
  title={Attention is all you need},
  author={Vaswani, Ashish and Shazeer, Noam and Parmar, Niki and Uszkoreit, Jakob and Jones, Llion and Gomez, Aidan N and Kaiser, {\L}ukasz and Polosukhin, Illia},
  journal={Advances in neural information processing systems},
  volume={30},
  year={2017}
}

@inproceedings{ronneberger2015unetconvolutionalnetworksbiomedical,
  title={U-net: Convolutional networks for biomedical image segmentation},
  author={Ronneberger, Olaf and Fischer, Philipp and Brox, Thomas},
  booktitle={International Conference on Medical image computing and computer-assisted intervention},
  pages={234--241},
  year={2015},
  organization={Springer}
}

@inproceedings{piga2024syntheticdatagenerationidentification,
  title={Synthetic data generation for system identification: leveraging knowledge transfer from similar systems},
  author={Piga, Dario and Rufolo, Matteo and Maroni, Gabriele and Mejari, Manas and Forgione, Marco},
  booktitle={2024 IEEE 63rd Conference on Decision and Control (CDC)},
  pages={6383--6388},
  year={2024},
  organization={IEEE}
}

@article{bazzi2024robomorphincontextmetalearningrobot,
  title={RoboMorph: In-context meta-learning for robot dynamics modeling},
  author={Bazzi, Manuel Bianchi and Shahid, Asad Ali and Agia, Christopher and Alora, John and Forgione, Marco and Piga, Dario and Braghin, Francesco and Pavone, Marco and Roveda, Loris},
  journal={International Conference on Informatics in Control, Automation and Robotics (ICINCO)},
  year={2025}
}

@article{ho2020denoisingdiffusionprobabilisticmodels,
  title={Denoising diffusion probabilistic models},
  author={Ho, Jonathan and Jain, Ajay and Abbeel, Pieter},
  journal={Advances in neural information processing systems},
  volume={33},
  pages={6840--6851},
  year={2020}
}

@inproceedings{perez2017filmvisualreasoninggeneral,
  title={Film: Visual reasoning with a general conditioning layer},
  author={Perez, Ethan and Strub, Florian and De Vries, Harm and Dumoulin, Vincent and Courville, Aaron},
  booktitle={Proceedings of the AAAI conference on artificial intelligence},
  volume={32},
  number={1},
  year={2018}
}

@ARTICLE{8772145,
  author={Gaz, Claudio and Cognetti, Marco and Oliva, Alexander and Robuffo Giordano, Paolo and De Luca, Alessandro},
  journal={IEEE Robotics and Automation Letters}, 
  title={Dynamic Identification of the Franka Emika Panda Robot With Retrieval of Feasible Parameters Using Penalty-Based Optimization}, 
  year={2019},
  volume={4},
  number={4},
  pages={4147-4154},
  doi={10.1109/LRA.2019.2931248}
}

@article{wang2023incontextlearningunlockeddiffusion,
  title={In-context learning unlocked for diffusion models},
  author={Wang, Zhendong and Jiang, Yifan and Lu, Yadong and He, Pengcheng and Chen, Weizhu and Wang, Zhangyang and Zhou, Mingyuan and others},
  journal={Advances in Neural Information Processing Systems},
  pages={8542--8562},
  year={2023}
}

@inproceedings{ramesh2023physicsinformedmodelbasedreinforcementlearning,
  title={Physics-informed model-based reinforcement learning},
  author={Ramesh, Adithya and Ravindran, Balaraman},
  booktitle={Learning for Dynamics and Control Conference},
  pages={26--37},
  year={2023},
  organization={PMLR}
}

@article{Ai2025LearningDynamicsReview,
  title={A review of learning-based dynamics models for robotic manipulation},
  author={Ai, Bo and Tian, Stephen and Shi, Haochen and Wang, Yixuan and Pfaff, Tobias and Tan, Cheston and Christensen, Henrik I and Su, Hao and Wu, Jiajun and Li, Yunzhu},
  journal={Science Robotics},
  volume={10},
  number={106},
  pages={eadt1497},
  year={2025},
  publisher={American Association for the Advancement of Science}
}

@article{vanschoren2018meta,
  title={Meta-learning: A survey},
  author={Vanschoren, Joaquin},
  journal={arXiv preprint arXiv:1810.03548},
  year={2018}
}

@article{giacomuzzos2023blackbox,
  title={A black-box physics-informed estimator based on gaussian process regression for robot inverse dynamics identification},
  author={Giacomuzzo, Giulio and Carli, Ruggero and Romeres, Diego and Dalla Libera, Alberto},
  journal={IEEE Transactions on Robotics},
  volume={40},
  pages={4820--4836},
  year={2024},
  publisher={IEEE}
}

@inproceedings{52580,
  title={Transformers learn in-context by gradient descent},
  author={Von Oswald, Johannes and Niklasson, Eyvind and Randazzo, Ettore and Sacramento, Jo{\~a}o and Mordvintsev, Alexander and Zhmoginov, Andrey and Vladymyrov, Max},
  booktitle={International Conference on Machine Learning},
  pages={35151--35174},
  year={2023},
  organization={PMLR}
}

@article{meng2024metapnn,
  title={Transferring Meta-Policy from Simulation to Reality via Progressive Neural Networks},
  author={Meng, Wei and Ju, Hao and Ai, Tongxu and Gomez, Randy and Nichols, Eric and Li, Guangliang},
  journal={IEEE Robotics and Automation Letters},
  year={2024}
}

@article{forgione2021dynonet,
  title={dynoNet: A neural network architecture for learning dynamical systems},
  author={Forgione, Marco and Piga, Dario},
  journal={International Journal of Adaptive Control and Signal Processing},
  volume={35},
  number={4},
  pages={612--626},
  year={2021},
  publisher={Wiley Online Library}
}

@inproceedings{hansen2024tdmpc2,
	title={TD-MPC2: Scalable, Robust World Models for Continuous Control}, 
	author={Nicklas Hansen and Hao Su and Xiaolong Wang},
	booktitle={International Conference on Learning Representations (ICLR)},
	year={2024}
}

@inproceedings{kirschgeneral,
  title={General-Purpose In-Context Learning by Meta-Learning Transformers},
  author={Kirsch, Louis and Harrison, James and Sohl-Dickstein, Jascha and Metz, Luke},
  booktitle={Sixth Workshop on Meta-Learning at the Conference on Neural Information Processing Systems},
  year = {2022}
}

@INPROCEEDINGS{paramidrobots1985,
  author={Khosla, Pradeep K. and Kanade, Takeo},
  booktitle={1985 24th IEEE Conference on Decision and Control}, 
  title={Parameter identification of robot dynamics}, 
  year={1985},
  volume={},
  number={},
  pages={1754-1760}}

@ARTICLE{5153127,
  author={Wang, Yang and Boyd, Stephen},
  journal={IEEE Transactions on Control Systems Technology}, 
  title={Fast Model Predictive Control Using Online Optimization}, 
  year={2010},
  volume={18},
  number={2},
  pages={267-278}}

@inproceedings{elseiagy2026data,
  title={Data-Driven Dynamic Parameter Learning of manipulator robots},
  author={Elseiagy, Mohammed and Alemayoh, Tsige Tadesse and Bezerra, Ranulfo and Kojima, Shotaro and Ohno, Kazunori},
  booktitle={2026 IEEE/SICE International Symposium on System Integration (SII)},
  pages={193--198},
  year={2026},
  organization={IEEE}
}

@inproceedings{
makoviychuk2021isaac,
title={Isaac Gym: High Performance {GPU} Based Physics Simulation For Robot Learning},
author={Viktor Makoviychuk and others},
booktitle={Thirty-fifth Conference on Neural Information Processing Systems Datasets and Benchmarks Track},
year={2021}
}

@article{Moroncelli2024TheDO,
  title={The duality of generative AI and reinforcement learning in robotics: A review},
  author={Angelo Moroncelli and Vishal Soni and Marco Forgione and Dario Piga and Blerina Spahiu and Loris Roveda},
  journal={Inf. Fusion},
  year={2024},
  volume={129},
  pages={104003}
}

\addtolength{\textheight}{0.6cm}   

\end{document}